\documentclass[11pt]{article}

\usepackage[final]{acl}

\usepackage{times}
\usepackage{latexsym}

\usepackage[T1]{fontenc}

\usepackage[utf8]{inputenc}

\usepackage{microtype}

\usepackage{inconsolata}

\usepackage{graphicx}

\usepackage{multirow}
\usepackage{xcolor}
\usepackage{colortbl}
\usepackage[table]{xcolor}
\usepackage{booktabs}
\usepackage{enumitem}
\usepackage{makecell}
\usepackage{amssymb}
\usepackage{amsmath}
\usepackage{bbm}
\usepackage{subcaption}

\definecolor{tablecolor}{HTML}{AABCDB}
\definecolor{ablationcolor}{HTML}{DACFE5}
\definecolor{mathcolor}{HTML}{FFF2DF}
\definecolor{codecolor}{HTML}{EAEBD7}
\definecolor{qacolor}{HTML}{F8D5D5}
\definecolor{trustcolor}{HTML}{B5C9DF}
\definecolor{robustcolor}{HTML}{D2C4B2}
\definecolor{auditorpromptcolor}{HTML}{E3B87F}
\definecolor{softpromptcolor}{HTML}{EDB8B0}
\definecolor{IMISpromptcolor}{HTML}{B2D3A4}

\usepackage[most]{tcolorbox}
\tcbuselibrary{skins, breakable}

\newtcolorbox{auditorpromptbox}[1]{
    breakable,
    colback=white,
    colframe=auditorpromptcolor,
    arc=4pt,
    boxrule=0.8pt,
    title=#1,
    coltitle=white,
    colbacktitle=auditorpromptcolor,
    fonttitle=\bfseries
}

\newtcolorbox{softpromptbox}[1]{
    breakable,
    colback=white,
    colframe=softpromptcolor,
    arc=4pt,
    boxrule=0.8pt,
    title=#1,
    coltitle=white,
    colbacktitle=softpromptcolor,
    fonttitle=\bfseries
}

\newtcolorbox{IMISpromptbox}[1]{
    breakable,
    colback=white,
    colframe=IMISpromptcolor,
    arc=4pt,
    boxrule=0.8pt,
    title=#1,
    coltitle=white,
    colbacktitle=IMISpromptcolor,
    fonttitle=\bfseries
}

\title{Belief-Guided Inference Control for Large Language Model Services via Verifiable Observations}

\author{
 \textbf{Wenhao Yuan\textsuperscript{1}},
 \textbf{Chenchen Lin\textsuperscript{2}},
 \textbf{Jian Chen\textsuperscript{1}},
 \textbf{Jinfeng Xu\textsuperscript{1}},
\\
 \textbf{Shuo Yang\textsuperscript{1}},
 \textbf{Edith Cheuk Han Ngai\textsuperscript{1,\thanks{Corresponding author.}}}
\\
 \textsuperscript{1}The University of Hong Kong,
 \textsuperscript{2}Sun Yat-sen University
\\
    \texttt{\href{mailto:wenhao.yuan@connect.hku.hk}{wenhao.yuan@connect.hku.hk}}, \texttt{\href{mailto:chngai@eee.hku.hk}{chngai@eee.hku.hk}}
}

\begin{document}
\maketitle
\begin{abstract}
In black-box large language model (LLM) services, response reliability is often only partially observable at decision time, while stronger inference pathways incur substantial computational cost, inducing a budgeted sequential decision problem: for each request, the system should decide whether the default low-cost response is sufficiently reliable or whether additional computation should be allocated to improve response quality. In this paper, we propose \textbf{Ver}ifiable \textbf{O}bservations for Risk-aware \textbf{I}nference \textbf{C}ontrol (\textsc{Veroic}), a framework for adaptive inference control in black-box LLM settings, which formulates request-time control as a \textit{partially observable Markov decision process} to capture partial observability and sequential budget coupling. It constructs a lightweight verifiable observation channel from the input-output pair by aggregating heterogeneous quality signals into a belief state over latent response reliability, which is then used by a budget-aware policy to decide whether to return the default output or trigger a higher-cost inference pathway. Experiments on diverse tasks show that \textsc{Veroic} achieves improved quality-cost trade-offs, stronger risk estimation and calibration, and more robust long-horizon inference control than competitive baselines.
\end{abstract}

\section{Introduction}
Large Language Models (LLMs) have become an important substrate for intelligent systems, supporting applications such as programming assistance~\citep{chen2025need, dinh2023large}, content creation~\citep{choi2025proxona, cheng2025beyond, padmakumar2024does}, and scientific reasoning~\citep{zhang2025scientific, lyu2025adapting, si2025can}. In practical deployments, however, LLMs are accessed as black-box services rather than as fully controlled models. Users and downstream systems observe only the input-output interface, while the reliability of each response is not directly observable at decision time and must instead be inferred from imperfect observable evidence~\citep{liang2023holistic, ouyang2022training}. Meanwhile, stronger inference pathways, such as larger models or longer reasoning, can improve response quality but incur additional latency and cost. 

Such black-box LLM services pose a practical online control problem. At request time, the system cannot directly observe whether the default low-cost response is reliable and must instead rely on imperfect response-side signals to decide whether additional computation should be allocated~\citep{kadavath2022language, kuhn2023semantic}. Stronger inference paths, such as escalation to a larger model, or iterative refinement, can improve response reliability, but they also incur nontrivial latency and monetary cost~\citep{dohan2022language, ong2024routellm}. Uniformly applying the strongest configuration is inefficient, while always accepting the default response risks failures. Moreover, the observable evidence available after the default response is often noisy and only partially informative, so myopic one-step routing may be insufficient. The objective is therefore not unconditional quality maximization, but adaptive control of the long-term quality--cost trade-off under partial observability.

Existing work has shown that LLM behavior can be variable and difficult to characterize from surface outputs alone~\citep{richter2025an, sun2025invisible, kemmerzell2025towards, balayn2025unpacking, yuan2026verify}. However, many existing approaches still evaluate or optimize LLM systems under the assumption that response quality can be inferred reliably either from observable outputs or from static offline benchmarks, which is violated in black-box online use: the response that appears superficially plausible may still be unreliable, while a conservative response may already be adequate for the task. This creates a gap between offline evaluation and the online control problem faced by real LLM services, giving rise to two core challenges. \textit{\textbf{\uppercase\expandafter{\romannumeral1}. Partial and noisy reliability observations:}} the true reliability of a response is usually not directly observable when the system must act. Lightweight signals extracted from the input-output pair, such as verifiable checks or proxy indicators, are informative but imperfect, and their reliability depends on context. \textit{\textbf{\uppercase\expandafter{\romannumeral2}. Budgeted long-horizon inference control:}} when stronger inference paths are costly, the system must repeatedly decide whether to allocate additional computation using limited evidence, balancing quality gains against inference cost under a budget constraint, leading to the question: \textit{How can a black-box LLM service maintain calibrated estimates of latent response reliability and allocate inference resources adaptively over time to optimize the long-term quality-cost trade-off?}

To address this problem, we propose \textbf{Ver}ifiable \textbf{O}bservations for Risk-aware \textbf{I}nference \textbf{C}ontrol (\textsc{Veroic}) in black-box LLM settings. Specifically, we formulate request-time inference management as a partially observable Markov decision process (POMDP) in \S~\ref{problem_formulation}, where the latent state captures the unobserved reliability condition of the default response at decision time. To address Challenge \textbf{\uppercase\expandafter{\romannumeral1}}, we construct calibrated lightweight reliability signals from the input-output pair and integrate them through the structured observation model and Bayesian belief updates in \S~\ref{verifiable_channel} to infer a belief state over latent response reliability. To address Challenge \textbf{\uppercase\expandafter{\romannumeral2}}, we use the belief state to parameterize a budget-aware control policy to decide whether to return the default response or escalate to a higher-cost inference path, enabling adaptive inference allocation under partial observability and sequential budget constraints. Unlike conventional routing methods, \textsc{Veroic} explicitly treats response reliability as a latent variable and performs inference control through belief-based decision making. Our main contributions are summarized as follows:
\begin{itemize}
\item We formulate the black-box LLM inference management as a partially observable sequential decision problem.

\item We propose \textsc{Veroic}, a novel belief-based framework that infers latent response reliability from verifiable checks and proxy signals for budget-aware escalation.

\item Experiments show that \textsc{Veroic} improves quality-cost trade-offs, risk estimation, calibration, and control robustness over baselines.
\end{itemize}




\section{Related Works}

\paragraph{Uncertainty Estimation and Calibration in LLMs.}
A growing body of work studies uncertainty estimation and calibration in large language models~\citep{xia2025survey, liu2025uncertainty, tao2025revisiting, xiong2024can}. Prior work examines whether LLMs can express calibrated confidence and how confidence signals align with actual correctness~\citep{xiong2024can, liu2024can, liu2025metafaith}. Related studies also investigate observable indicators of unreliability, such as hallucination, inconsistency, and unsupported claims, often with auxiliary evaluators or self-consistency signals~\citep{sriramanan2024llm, du2024haloscope, simhi2025trust}. However, these methods mostly operate at the level of individual outputs and do not address sequential, budget-aware inference allocation.

\paragraph{Adaptive Inference Routing and Cost-Aware LLM Systems.}
Recent work explores routing and cost-aware control strategies for improving the quality-cost trade-off of LLM systems. \textsc{FrugalGPT} routes requests across models with different cost-quality operating points~\citep{chen2024frugalgpt}, while more recent work studies adaptive routing based on certainty or confidence signals~\citep{lu2025prolonged}. Related approaches use observable output features, uncertainty scores, or lightweight verifiers to decide when stronger inference is needed~\citep{lu2025prolonged, chen2024frugalgpt}. Most such methods map current observable features directly to one-step decisions, without explicitly modeling latent response reliability or maintaining a belief state.

\paragraph{Decision Making under Partial Observability.}
A related line of work highlights that observable outputs alone are often insufficient for reliably characterizing LLM behavior, especially under uncertainty, hidden service variation, or imperfect monitoring~\citep{liang2023holistic, sun2025invisible}. Other studies show that evaluation or monitoring signals can be incomplete or misleading, making one-step decisions based only on surface observations unreliable~\citep{denison2024sycophancy, ji2025mitigating, wang2025quality}. In contrast to static evaluation or episodic monitoring protocols, our setting focuses on sequential decision making under partial observability, where inference actions are coupled through a shared computation budget. \textsc{Veroic} builds on this perspective by maintaining a posterior belief over latent response reliability and using it to guide budget-aware escalation over time.

\begin{figure*}[t]
\centerline{\includegraphics[width=0.93\textwidth, trim=0 0 0 0,clip]{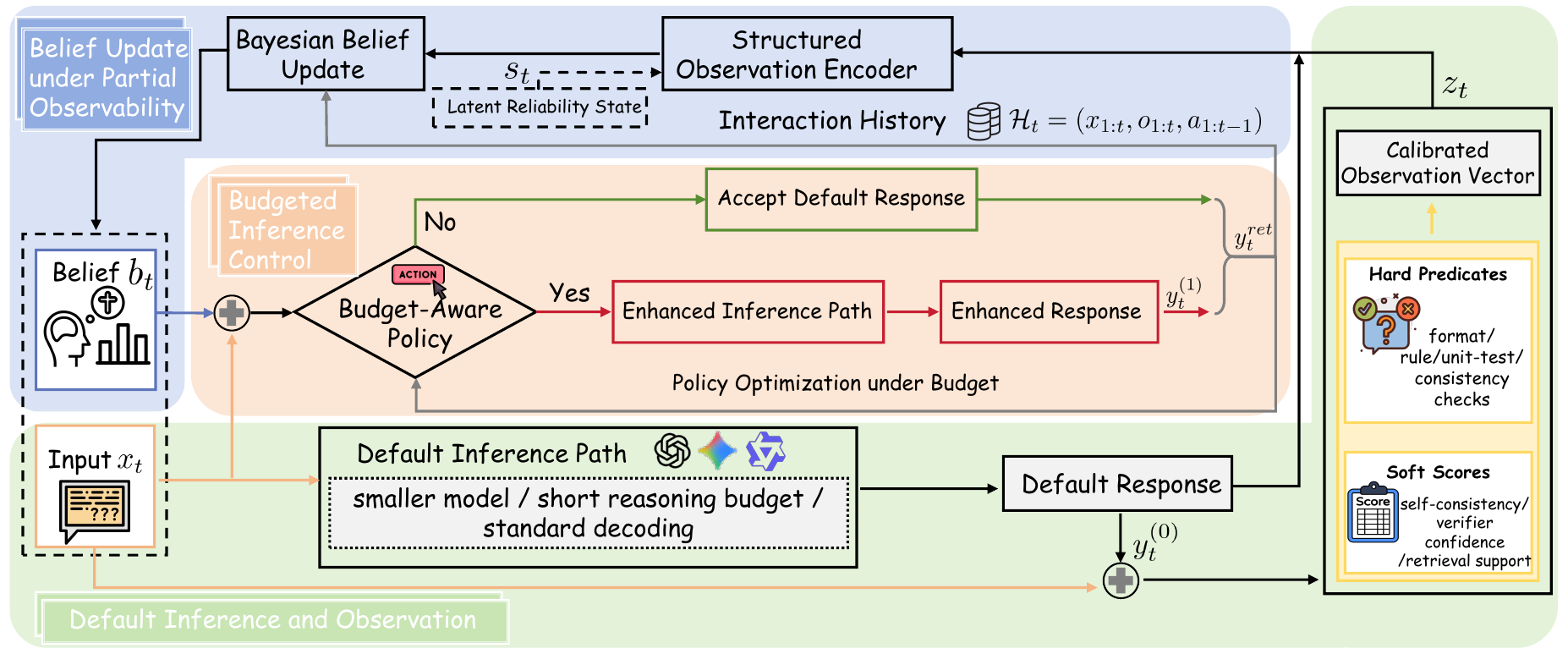}}
\caption{The framework overview of our proposed \textsc{Veroic}.}
\label{framework}
\vspace{-10pt}
\end{figure*}

\section{Problem Formulation} \label{problem_formulation}
We study adaptive inference control in black-box LLM settings. At each step $t$, the system receives a request $x_t \in \mathcal{X}$ and first produces a default low-cost response $y_t^{(0)} \in \mathcal{Y}$, such as one generated by a smaller model or a shorter reasoning budget. While the default response is inexpensive, its true reliability is generally not directly observable at decision time. A stronger inference path can improve response reliability, but incurs additional cost. We consider an online setting in which requests arrive sequentially and the enhanced inference budget is shared across the interaction horizon, so earlier decisions affect later computation allocation. The core problem is therefore to decide, for each request, whether to accept the default response or allocate additional inference computation.

We formulate the problem as a \textit{partially observable Markov decision process}, denoted by $\mathcal{M}=(\mathcal{S}, \mathcal{A},\Omega,\mathcal{T},\mathcal{O},\mathcal{R},\gamma)$. The latent state $s_t \in \mathcal{S}$ captures the unobserved reliability condition of the default response for the current request and is influenced by factors that are not directly visible to the controller, including request difficulty, inference uncertainty, and other hidden factors affecting response reliability. We instantiate the latent state space as a binary set $\mathcal{S} = \{0,1\}$, where $s_t=1$ indicates that the default response is reliable and $s_t=0$ otherwise. Since the latent state is not directly observable, the controller maintains a belief state $b_t(s)=\mathbb{P}(s_t=s\mid \mathcal{H}_t)$, where $\mathcal{H}_t$ denotes the interaction history up to step $t$. The belief state summarizes the currently available evidence about latent response reliability and therefore serves as the decision state under partial observability. The action space $\mathcal{A}=\{0,1\}$, where $a_t=0$ means that the system accepts the default response and returns $y_t^{(0)}$ as the final output, while $a_t=1$ means that the system triggers an enhanced inference path and obtains an alternative response $y_t^{(1)}$, which may be produced by a stronger model or a larger reasoning budget. The observation space is $\Omega := \mathcal{Y}\times (0,1)^k$, and the observation at step $t$ is written as $o_t = (y_t^{(0)}, z_t)$, where $y_t^{(0)}$ is the default response and $z_t \in (0,1)^k$ is a lightweight quality observation vector constructed from $(x_t, y_t^{(0)})$. After generating the default response, the controller extracts the lightweight observation vector and updates its belief over the latent state. The control policy is then defined by $\pi(a_t \mid b_t, x_t)$, where the belief state summarizes the current observation together with the interaction history, and $x_t$ is retained to preserve request-specific contextual information not fully captured by the low-dimensional belief state. The transition kernel $\mathcal{T}(s_{t+1}\mid s_t, x_t, a_t)$ models the sequential evolution of latent reliability conditions across requests and can capture service-side non-stationarity when such effects are present. The observation kernel $\mathcal{O}(o_t \mid s_t, x_t)$ connects the latent state to the observable default response and quality signals. We will specify the observation in \S~\ref{verifiable_channel}.

Within the POMDP formulation, the controller uses the current belief state to decide whether to accept the default response or invoke enhanced inference. The controller seeks a policy that optimizes long-term utility under a computation budget. The reward trades off the quality of the response against the additional cost of enhanced inference, while optionally penalizing decisions made under high estimated risk. The resulting objective is to allocate stronger inference selectively, rather than uniformly, so as to optimize the long-term quality-cost trade-off under partial observability.

\section{Belief-Based Inference Control} \label{verifiable_channel}

\subsection{Lightweight Observation Construction}
To achieve belief-based inference control without incurring the cost of stronger inference on every request, we construct a lightweight quality observation channel from the input-output pair $(x_t, y_t^{(0)})$. The goal is to approximate the latent quality condition of the default response using low-cost but informative signals that are available immediately after the default inference pass. Since no single signal is sufficiently reliable across tasks and requests, we combine heterogeneous evidence sources into a unified observation vector for downstream belief updating. (\expandafter{\romannumeral1}) \emph{Hard verifiable checks} $h_j$ are Boolean predicates that capture discrete properties that can be programmatically evaluated, such as format validity, rule satisfaction, or unit-test results. Their instantaneous realizations are defined as
\begin{align}
\tilde{h}_{t}^{(j)} = \mathbbm{1}\{h_j(x_t, y_t^{(0)})\}, j\in[k_h].
\end{align}
(\expandafter{\romannumeral2}) \emph{Soft proxy scores} $g_i$ are bounded score functions that provide continuous quality-related evidence, such as self-consistency, verifier confidence, or retrieval support, which produce the raw scores
\begin{align}
\tilde{g}_{t}^{(i)} = g_i(x_t, y_t^{(0)}), i\in[k_g].
\end{align}

Due to the heterogeneous and noisy signals, we further smooth and calibrate them before belief updating. For hard signals, we maintain a sliding-window summary over the most recent $W$ realizations to reduce sensitivity to one-step noise and to capture short-term temporal consistency. For soft scores, we apply a monotone calibration map so that different proxy dimensions become comparable as probabilistic indicators of latent response reliability. Let $\Xi_t^{(j)} =\sum_{\ell=\max(1,t-W+1)}^{t}\tilde{h}_{\ell}^{(j)}$ denote the windowed count for the $j$-th hard signal. We then define the components as
\begin{align}
z_{h,t}^{(j)} = \frac{\alpha_{0j}+\Xi_t^{(j)}}{\alpha_{0j}+\beta_{0j}+W}, z_{g,t}^{(i)} = \Gamma_i(\tilde{g}_t^{(i)};\vartheta_{i,t}),
\end{align}
where $(\alpha_{0j},\beta_{0j})$ are Beta-Bernoulli pseudocounts, and $\Gamma_i$ is the calibration function with parameters $\vartheta_{i,t}$. By collecting all the calibrated components, it yields the observation vector
\begin{align}
z_t = (z_{h,t}^{(1:k_h)}, z_{g,t}^{(1:k_g)}) \in (0,1)^k, k \!=\! k_h + k_g,
\end{align}
which serves as a lightweight observation of the latent quality state associated with the default response. Importantly, $z_t$ is not treated as a direct quality label; rather, it provides noisy but continuous evidence for posterior belief updating under partial observability, enabling the controller to estimate response reliability and determine whether enhanced inference is warranted.

\subsection{Structured Observation Model}
Given $x_t$, the system first produces $y_t^{(0)}$ and then constructs the full observation tuple $o_t = (y_t^{(0)}, z_t) \in \Omega$. As the true response-generation mechanism of a black-box LLM service is inaccessible, we introduce a structured observation encoder that maps the default response and proxy signals into state-dependent evidence for posterior belief updating. Let $\psi_t:= (\{\Xi_t^{(j)}\}_{j=1}^{k_h}, \{\vartheta_{i,t}\}_{i=1}^{k_g})$ denote the observation-side calibration state associated with the hard and soft signals. We define a state-conditioned compatibility score
\begin{align}
\!\!\! \ell_\phi(s_t; x_t, y_t^{(0)}, z_t, \psi_t) = &\ \ell_y(s_t; x_t, y_t^{(0)}) \nonumber \\
+ & \ell_z(s_t; x_t, y_t^{(0)}, z_t, \psi_t),
\end{align}
where $\ell_y$ captures semantic evidence extracted from the default response, and $\ell_z$ captures auxiliary reliability evidence provided by the calibrated observation vector. This decomposition separates information contained in the surface response from information provided by external quality proxies, which is important because a plausible-looking response may still be unreliable, while lightweight proxy signals may reveal latent reliability risk not directly visible from the text alone. Within $\ell_z$, the hard components $z_{h,t}$ are treated as deterministic calibrated summaries induced by the sliding-window calibration state $\psi_t$, whereas the soft components $z_{g,t}$ are processed through state-dependent compatibility functions. We use the factorized form
\begin{align}
\!\!\! \ell_z(s_t; x_t, y_t^{(0)}\!, z_t, \psi_t) \!=\! &\ \ell_h(x_t, y_t^{(0)}, z_{h,t}; \psi_t) \nonumber \\
+\! \sum\nolimits_{i=1}^{k_g}\ & \ell_g^{(i)}(s_t; x_t, y_t^{(0)}\!, z_{g,t}^{(i)}; \psi_t),\!\!
\end{align}
where $\ell_h$ summarizes the contribution of the calibrated hard evidence and $\ell_g^{(i)}$ measures how compatible the $i$-th soft proxy dimension is with the latent reliability state. $\ell_y$, $\ell_h$, and $\ell_g^{(i)}$ can be instantiated by lightweight parametric scoring modules over response features and calibrated proxy inputs. Next, we define the observation likelihood up to normalization as $\mathcal{O}(o_t \!\mid\! s_t, x_t; \psi_t) \!\propto\! \exp (\ell_\phi(s_t; x_t, y_t^{(0)}\!, z_t, \psi_t))$, which provides a learnable interface that converts response content and calibrated proxy signals into posterior evidence over the latent state. In practice, the encoder parameters $\phi$ are fitted using sparse verification feedback, delayed task-level supervision, or off-policy trajectories collected during policy optimization. As a result, the observation model remains directly implementable in black-box online settings while preserving the state-based structure needed for belief updating and downstream inference control.

\subsection{Bayesian Belief Dynamics}
Since the true quality of the response is unobservable, the controller maintains a Bayesian belief over latent reliability states throughout the online interaction process. Given the history up to step $t$, $\mathcal{H}_t = (x_{1:t}, o_{1:t}, a_{1:t-1})$, the belief state is defined as $b_t(s) = \mathbb{P}(s_t = s \mid \mathcal{H}_t)$, $s \in \mathcal{S}$, which summarizes the currently available evidence about latent response reliability and serves as the sufficient decision state under partial observability. At each step $t$, the system receives the request $x_t$, executes the default low-cost inference path, and constructs the observation tuple $o_t$. The controller then updates its belief through an approximate Bayesian prediction-filtering procedure. First, the previous posterior is propagated through the latent transition model to obtain the predictive belief $\hat{b}_t(s_t) \!=\! \sum_{s_{t-1}\in\mathcal{S}} \mathcal{T}(s_t \mid s_{t-1}, x_{t-1}, a_{t-1}) b_{t-1}(s_{t-1})$, capturing temporal persistence in latent reliability conditions induced by the online deployment process. Once observing the response and quality observation vector, the controller performs Bayesian filtering with the structured observation encoder:
\begin{align}\label{bayes}
b_t(s_t) = \frac{\mathcal{O}(o_t \mid s_t, x_t; \psi_t)\,\hat{b}_t(s_t)}{\sum_{s' \in \mathcal{S}}\mathcal{O}(o_t \mid s', x_t; \psi_t)\,\hat{b}_t(s')},
\end{align}
which combines two complementary sources of information and shows how the observation encoder transforms the default response and proxy evidence into posterior mass over the latent reliability state. The default response $y_t^{(0)}$ provides semantic evidence about the quality of the current output, while $z_t$ contributes lightweight but informative proxy signals about latent reliability risk. Their combination refines the predictive belief into a posterior estimate that is more stable than any single noisy signal alone, thereby turning imperfect per-request observations into a decision state suitable for long-horizon, budget-aware inference control.

\subsection{Budgeted Risk-Aware Policy Optimization}
Based on the current belief state, the controller decides whether to accept the default response or invoke enhanced inference. We optimize this policy under a long-term computation budget, with the goal of balancing realized task utility, inference cost, and residual reliability risk under partial observability. Let $y_t^{\mathrm{ret}}$ denote the final returned response, which equals the default output $y_t^{(0)}$ when $a_t=0$ and the enhanced output $y_t^{(1)}$ when $a_t=1$. We define the per-step reward as
\begin{align}
r_t = V(y_t^{\mathrm{ret}}) - C(a_t) - \kappa R(b_t),
\end{align}
where $V(\cdot)$ denotes an observable utility surrogate instantiated from task-specific feedback, evaluation signals, or delayed supervision available during training or policy optimization, $C(a_t)$ denotes the additional computation cost induced by enhanced inference, and $\kappa \ge 0$ controls the penalty placed on residual reliability risk at decision time. The belief-dependent risk term is defined as $R(b_t):= \sum_{s\in\mathcal{S}} b_t(s)\varrho(s)$, where $\varrho(s) \geq 0$ is a state-dependent risk score reflecting the unreliability associated with latent state $s$. To prevent excessive escalation to high-cost inference, we impose a discounted computation budget $\mathbb{E}_{\pi}[\sum_{t\ge0}\gamma^t C(a_t)] \leq B$, where discount factor $\gamma\in(0,1)$ and $B$ is the effective long-run budget. Equivalently, the constraint can be written as $\mathbb{E}_{\pi}[\sum_{t\ge0}\gamma^t (B(1-\gamma)-C(a_t))]\ge 0$, which enables a Lagrangian formulation of the constrained control objective. To encourage sufficient exploration during policy learning, we further incorporate a Shannon entropy regularizer $\mathcal{H}(\pi)$, which mitigates premature convergence to degenerate escalation policies under uncertain beliefs. The resulting optimization objective is
\begin{align}
\max_{\pi}  \mathbb{E}_{\pi}[\sum\nolimits_{t\ge0} &\ \gamma^t (r_t + \mu_d(B(1-\gamma)-C(a_t)) \nonumber \\
&+ \rho \mathcal{H}(\pi(\cdot\mid b_t,x_t)))],
\end{align}
where $\mu_d \ge 0$ is the dual variable associated with the computation-budget constraint and $\rho \ge 0$ is the entropy weight. The entropy term is given by $\mathcal{H}(\pi(\cdot\mid b_t,x_t)) = -\sum_{a\in\mathcal{A}} \pi(a\mid b_t,x_t)\log \pi(a\mid b_t,x_t)$. In practice, we optimize the belief-conditioned policy with actor-critic style policy gradient updates under the Lagrangian objective. The objective encourages the policy to allocate stronger inference only when the expected quality gain justifies the additional computation cost. When the posterior belief indicates that the default response is likely reliable, the controller is encouraged to preserve computation by accepting the low-cost output; when the belief places substantial mass on high-risk states, the controller is encouraged to trigger enhanced inference despite its additional cost. In this way, the learned policy achieves adaptive long-horizon control of the quality-cost trade-off under partial observability.

\begin{table*}[t]
\centering
\caption{Performance of \textsc{Veroic} and other baselines on various tasks (\setlength{\fboxsep}{1pt}\colorbox{mathcolor}{Math Reasoning}, \colorbox{codecolor}{Code Generation}, and \colorbox{qacolor}{Question Answering}) under the CMIS setting. The best result in each column is marked in \textbf{bold}.}
\renewcommand\arraystretch{0.9}
\resizebox{1.0\textwidth}{!}{
\begin{tabular}
{c|c|>{\columncolor{mathcolor}}c|>{\columncolor{mathcolor}}c|>{\columncolor{codecolor}}c|>{\columncolor{codecolor}}c|>{\columncolor{qacolor}}c >{\columncolor{qacolor}}c|>{\columncolor{qacolor}}c >{\columncolor{qacolor}}c|>{\columncolor{qacolor}}c >{\columncolor{qacolor}}c} 
\toprule[1.2pt]
\multirow{2}{*}{\multirowcell{1.5}{\centering\textbf{Inference Paths}}} & \multirow{2}{*}{\multirowcell{1.5}{\centering\textbf{Methods}}} & \multicolumn{1}{>{\columncolor{mathcolor}}c|}{\centering\textbf{GSM8K}}  & \multicolumn{1}{>{\columncolor{mathcolor}}c|}{\centering\textbf{MATH}}  & \multicolumn{1}{>{\columncolor{codecolor}}c|}{\centering\textbf{HumanEval}}  & \multicolumn{1}{>{\columncolor{codecolor}}c|}{\centering\textbf{MBPP}}  & \multicolumn{2}{>{\columncolor{qacolor}}c|}{\centering\textbf{HotpotQA}} & \multicolumn{2}{>{\columncolor{qacolor}}c|}{\centering\textbf{2WikiMHQA}} & \multicolumn{2}{>{\columncolor{qacolor}}c}{\centering\textbf{PopQA}} \\ \cmidrule[0.5pt](l{1pt}r{0pt}){3-12}

& & EM $\uparrow$ & EM $\uparrow$ & Pass@1 $\uparrow$ & Pass@1 $\uparrow$ & EM $\uparrow$ & F1 $\uparrow$ & EM $\uparrow$ & F1 $\uparrow$ & EM $\uparrow$ & F1 $\uparrow$ \\ \cmidrule[0.8pt](l{1pt}r{0pt}){1-12}

\multirow{5}{*}{\multirowcell{-1}{LLaMA-3.1-8B \\ $\downarrow$ \\ LLaMA-3.1-70B}}
        & \textsc{Fixed} & 86.8 & 45.8 & 65.5 & 67.2 & 51.6 & 68.7 & 41.2 & 57.6 & 31.1 & 47.4  \\ 

        & \textsc{Random} & 88.9 & 49.4 & 69.7 & 70.3 & 53.8 & 70.6 & 43.4 & 59.7 & 33.5 & 49.2  \\ 

        & \textsc{Periodic} & 89.4 & 49.8 & 69.3 & 70.6 & 54.7 & 70.8 & 43.9 & 60.3 & 33.2 & 49.6  \\ 

        & \textsc{Output-trigger} & 90.1 & 51.2 & 71.3 & 71.8 & 55.3 & 71.6 & 44.7 & 61.4 & 34.2 & 50.2   \\ 

        & \textsc{Confidence-Threshold} & 90.4 & 51.5 & 71.6 & 72.2 & 55.6 & 71.9 & 44.8 & 61.6 & 34.2 & 50.7    \\ 

        & \textsc{Risk Predictor} & 91.2 & 52.1 & 72.3 & 72.6 & 55.9 & 72.3 & 45.3 & 62.0 & 34.9 & 51.0   \\ 
        
        & \textsc{Veroic} & \textbf{92.4} & \textbf{52.7} & \textbf{72.9} & \textbf{73.2} & \textbf{56.2} & \textbf{72.8} & \textbf{45.8} & \textbf{62.6} & \textbf{35.3} & \textbf{51.5}   \\ \midrule[0.8pt]

\multirow{5}{*}{\multirowcell{-1}{Qwen-2.5-7B \\ $\downarrow$ \\ Qwen-2.5-32B}}
        & \textsc{Fixed} & 85.6 & 44.8 & 62.8 & 66.2 & 51.2 & 68.3 & 40.5 & 56.9 & 30.2 & 46.8  \\ 

        & \textsc{Random} & 87.9 & 48.5 & 66.4 & 69.6 & 53.1 & 70.2 & 42.8 & 59.2 & 32.5 & 48.7  \\ 

        & \textsc{Periodic} & 88.1 & 48.9 & 66.8 & 70.1 & 53.3 & 70.3 & 43.1 & 59.5 & 32.6 & 48.9  \\ 

        & \textsc{Output-trigger} & 89.4 & 50.6 & 68.7 & 71.8 & 54.6 & 71.3 & 44.5 & 60.8 & 33.7 & 49.8   \\ 
        
        & \textsc{Confidence-Threshold} & 89.7 & 50.9 & 68.9 & 72.0 & 54.9 & 71.6 & 44.7 & 61.0 & 33.9 & 50.1   \\ 

        & \textsc{Risk Predictor} & 90.6 & 51.8 & 70.1 & 72.3 & 55.3 & 71.9 & 45.1 & 61.8 & 34.3 & 50.5  \\ 
        
        & \textsc{Veroic} & \textbf{91.4} & \textbf{52.3} & \textbf{71.2} & \textbf{72.5} & \textbf{55.8} & \textbf{72.1} & \textbf{45.8} & \textbf{62.3} & \textbf{34.8} & \textbf{50.8}   \\ 

\bottomrule[1.2pt]
\end{tabular}}
\label{overall_performance_cmis}
\vspace{-5pt}
\end{table*}

\begin{table*}[ht]
\centering
\setlength{\tabcolsep}{3pt}
\caption{Risk estimation and probability calibration performance for identifying high-risk or low-quality outputs that may benefit from enhanced inference under the CMIS setting.}
\resizebox{1.0\textwidth}{!}{
\begin{tabular}{c|ccccc|ccccc|ccccc} 
\toprule[1.2pt]
\multirow{2}{*}{\multirowcell{2}{\centering\textbf{Methods}}} & \multicolumn{5}{c|}{\centering\textbf{GSM8K}} & \multicolumn{5}{c|}{\centering\textbf{HumanEval}} & \multicolumn{5}{c}{\centering\textbf{HotpotQA}} \\ \cmidrule[0.5pt](l{1pt}r{0pt}){2-16}

& AUROC $\uparrow$ & AUPRC $\uparrow$ & Brier $\downarrow$ & NLL $\downarrow$ & ECE $\downarrow$ & AUROC $\uparrow$ & AUPRC $\uparrow$ & Brier $\downarrow$ & NLL $\downarrow$ & ECE $\downarrow$ & AUROC $\uparrow$ & AUPRC $\uparrow$ & Brier $\downarrow$ & NLL $\downarrow$ & ECE $\downarrow$ \\ \cmidrule[0.8pt](l{1pt}r{0pt}){1-16}

\textsc{Fixed} & 0.71 & 0.32 & 0.185 & 0.523 & 0.085 & 0.66 & 0.26 & 0.205 & 0.584 & 0.096 & 0.64 & 0.23 & 0.218 & 0.615 & 0.114  \\ 

\textsc{Random} & 0.73 & 0.41 & 0.173 & 0.475 & 0.076 & 0.72 & 0.33 & 0.197 & 0.529 & 0.084 & 0.69 & 0.28 & 0.192 & 0.563 & 0.093  \\ 

\textsc{Periodic} & 0.77 & 0.43 & 0.162 & 0.467 & 0.065 & 0.74 & 0.38 & 0.182 & 0.522 & 0.076 & 0.72 & 0.31 & 0.185 & 0.548 & 0.085  \\ 

\textsc{Output-trigger} & 0.83 & 0.48 & 0.149 & 0.447 & 0.062 & 0.77 & 0.41 & 0.171 & 0.495 & 0.072 & 0.77 & 0.38 & 0.180 & 0.528 & 0.079  \\ 

\textsc{Confidence-Threshold} & 0.84 & 0.49 & 0.152 & 0.452 & 0.066 & 0.78 & 0.42 & 0.175 & 0.505 & 0.074 & 0.78 & 0.39 & 0.183 & 0.535 & 0.082  \\ 

\textsc{Risk Predictor} & 0.85 & 0.51 & 0.144 & 0.432 & 0.058 & 0.80 & 0.46 & 0.166 & 0.482 & 0.065 & 0.80 & 0.41 & 0.172 & 0.520 & 0.072   \\ 
        
\cellcolor{trustcolor}\textsc{Veroic} & \cellcolor{trustcolor}0.87 & \cellcolor{trustcolor}0.52 & \cellcolor{trustcolor}0.139 & \cellcolor{trustcolor}0.417 & \cellcolor{trustcolor}0.048 & \cellcolor{trustcolor}0.81 & \cellcolor{trustcolor}0.48 & \cellcolor{trustcolor}0.160 & \cellcolor{trustcolor}0.473 & \cellcolor{trustcolor}0.057 & \cellcolor{trustcolor}0.81 & \cellcolor{trustcolor}0.43 & \cellcolor{trustcolor}0.166 & \cellcolor{trustcolor}0.513 & \cellcolor{trustcolor}0.064  \\ 
        
\bottomrule[1.2pt]
\end{tabular}}
\label{degradation_detection_CMIS_LLaMA}
\vspace{-6pt}
\end{table*}

\section{Experiments}

\subsection{Experimental Setups}

\paragraph{Datasets.}
We evaluate our method on benchmark datasets covering diverse reasoning paradigms. \textbf{Math Reasoning} involves solving problems that require multi-step symbolic manipulation and logical deduction. We adopt GSM8K~\citep{cobbe2021training} and MATH~\citep{hendrycks2measuring} for this category. \textbf{Code Generation} evaluates the ability to synthesize executable programs from natural language specifications, and we consider HumanEval~\citep{chen2021evaluating} and MBPP~\citep{austin2021program}. \textbf{Question Answering} (QA) requires combining factual knowledge with reasoning over one or multiple pieces of evidence, testing both inference accuracy and factual consistency. We evaluate on HotpotQA~\citep{yang2018hotpotqa}, 2WikiMHQA~\citep{ho2020constructing}, and PopQA~\citep{mallen2023popqa}.

\paragraph{Baselines.}
We compare our method against several representative baselines. \textsc{Fixed}, \textsc{Random}, and \textsc{Periodic} are budget-matched heuristic strategies. \textsc{Output-Trigger}~\citep{lu2025prolonged, chen2024frugalgpt} is a rule-based router that escalates on predefined output-side signals, such as failed checks. \textsc{Confidence-Threshold}~\citep{xiong2024can, tao2025revisiting} escalates when a single confidence or risk score crosses a budget-calibrated threshold. \textsc{Risk Predictor}~\citep{liu2024can, liu2025uncertainty} trains a supervised model on the observation features to predict default-response failure risk, and escalates when the predicted risk exceeds a budget-calibrated threshold.

\paragraph{Evaluation Metrics.}
We evaluate the method from three complementary perspectives. \textbf{Task Utility}: For reasoning and QA tasks, we report EM and F1 scores; for code generation tasks, we report Pass@1. \textbf{Risk Estimation and Calibration}: We report AUROC and AUPRC to measure the ability of the signals to identify low-quality or high-risk outputs, and evaluate probabilistic calibration using the Brier score, negative log-likelihood (NLL), and expected calibration error (ECE). \textbf{Long-Horizon Control Performance}: To assess robustness, we report the low-quality occurrence rate (Occ), the conditional value-at-risk (CVaR) to capture tail risk, and the recovery delay (RecD), which measures recovery latency after severe low-quality events.

\paragraph{Inference Settings.}
We consider two inference settings to evaluate adaptive inference control. (\romannumeral1) \textit{Cross-Model Inference Switching} (CMIS). The system uses a lower-cost model as the default inference path and a stronger model as the enhanced inference path (e.g., LLaMA-3.1-8B $\rightarrow$ 70B). (\romannumeral2) \textit{Intra-Model Inference Scaling} (IMIS). The underlying base model remains unchanged, but the system varies the effective inference budget within the same model, such as reasoning depth, decoding budget, or refinement strength, yielding different quality-cost operating points. 

\paragraph{Implementation Details.}
We evaluate the controller under a sequential interaction protocol with a fixed episode length $H=500$. For each dataset, we uniformly sample 500 instances without replacement to form one episode and reveal them one step at a time. At each step, the controller has access only to the current request, the default response, and the interaction history up to that point, but not to future requests. Unless otherwise specified, results are reported under the CMIS setting with two backbone pairs: LLaMA-3.1-8B $\rightarrow$ LLaMA-3.1-70B and Qwen-2.5-7B $\rightarrow$ Qwen-2.5-32B, where the smaller model serves as the default low-cost inference path, and the larger model serves as the enhanced inference path.  We set the enhanced inference budget to $\alpha=0.05$ and the discount factor to $\gamma=0.99$. In budget-sensitivity experiments, we sweep $\alpha \in \{0.01, 0.02, 0.05, 0.10, 0.20\}$. We set the sliding-window length to $W=20$. The enhanced inference action is implemented using a higher-quality inference pipeline, instantiated here by a higher-capacity model or a larger reasoning budget. We instantiate $k_h=4$ (format and constraint checks, consistency or unit-test style checks, answerability or groundedness heuristics, and safety or compliance checks) with $k_g=2$ (self-consistency and an external lightweight verifier). The inference control policy is trained using an on-policy reinforcement learning algorithm and is frozen during evaluation. We set the enhanced inference cost to $C=0.20$, the risk penalty weight to $\kappa=1.0$, the entropy weight to $\rho=0.01$, and the learning rate to $\eta=0.05$. 

\begin{table}[t]
\centering
\setlength{\tabcolsep}{1pt}
\caption{Long-horizon control performance with the LLaMA inference-path pair under the CMIS setting.}
\resizebox{0.48\textwidth}{!}{
\begin{tabular}{c|ccc|ccc|ccc} 
\toprule[1.2pt]
\multirow{2}{*}{\multirowcell{2}{\centering\textbf{Methods}}} & \multicolumn{3}{c|}{\centering\textbf{GSM8K}} & \multicolumn{3}{c|}{\centering\textbf{HumanEval}} & \multicolumn{3}{c}{\centering\textbf{HotpotQA}} \\ \cmidrule[0.5pt](l{1pt}r{0pt}){2-10}

& Occ $\downarrow$ & CVaR $\downarrow$ & RecD $\downarrow$ & Occ $\downarrow$ & CVaR $\downarrow$ & RecD $\downarrow$ & Occ $\downarrow$ & CVaR $\downarrow$ & RecD $\downarrow$ \\ \cmidrule[0.8pt](l{1pt}r{0pt}){1-10}

\textsc{Fixed} & 0.42 & 0.38 & 210 & 0.48 & 0.41 & 243 & 0.52 & 0.48 & 276    \\ 

\textsc{Random} & 0.33 & 0.26 & 145 & 0.39 & 0.34 & 187 & 0.45 & 0.37 & 216    \\ 

\textsc{Periodic} & 0.25 & 0.22 & 134 & 0.34 & 0.28 & 166 & 0.39 & 0.30 & 193    \\ 

\textsc{Output-trigger} & 0.20 & 0.17 & 109 & 0.26 & 0.22 & 145 & 0.31 & 0.27 & 168   \\ 

\textsc{Confidence-Threshold} & 0.19 & 0.16 & 112 & 0.25 & 0.21 & 150 & 0.31 & 0.26 & 174   \\ 

\textsc{Risk Predictor} & 0.17 & 0.14 & 98 & 0.23 & 0.19 & 133 & 0.27 & 0.23 & 158    \\ 

\cellcolor{robustcolor}\textsc{Veroic} & \cellcolor{robustcolor}0.13 & \cellcolor{robustcolor}0.11 & \cellcolor{robustcolor}77 & \cellcolor{robustcolor}0.19 & \cellcolor{robustcolor}0.17 & \cellcolor{robustcolor}114 & \cellcolor{robustcolor}0.22 & \cellcolor{robustcolor}0.19 & \cellcolor{robustcolor}128    \\ 
        
\bottomrule[1.2pt]
\end{tabular}}
\label{stability_risk_CMIS_LLaMA}
\vspace{-12pt}
\end{table}

\begin{table*}[t]
\centering
\caption{Ablation study of our proposed \textsc{Veroic} under the CMIS setting.}
\resizebox{1.0\textwidth}{!}{
\begin{tabular}{c|cccc|cccc|ccccc} 
\toprule[1.2pt]
\multirow{2}{*}{\multirowcell{2}{\centering\textbf{Methods}}} & \multicolumn{4}{c|}{\centering\textbf{GSM8K}} & \multicolumn{4}{c|}{\centering\textbf{HumanEval}} & \multicolumn{5}{c}{\centering\textbf{HotpotQA}} \\ \cmidrule[0.5pt](l{1pt}r{0pt}){2-14}

& EM $\uparrow$ & Occ $\downarrow$ & CVaR $\downarrow$ & RecD $\downarrow$ & Pass@1 $\uparrow$ & Occ $\downarrow$ & CVaR $\downarrow$ & RecD $\downarrow$ & EM $\uparrow$ & F1 $\uparrow$ & Occ $\downarrow$ & CVaR $\downarrow$ & RecD $\downarrow$  \\ \cmidrule[0.8pt](l{1pt}r{0pt}){1-14}

w/o Risk Signals & 91.2 & 0.23 & 0.18 & 106 & 71.2 & 0.31 & 0.25 & 169 & 54.3 & 71.8 & 0.29 & 0.25 & 158  \\ 

w/o Enhanced Inference & 88.7 & 0.48 & 0.42 & 239 & 66.9 & 0.55 & 0.48 & 275 & 52.5 & 68.5 & 0.52 & 0.47 & 263  \\ 

w/o Adaptive Policy  & 90.4 & 0.30 & 0.24 & 142 & 69.4 & 0.29 & 0.25 & 187 & 54.1 & 70.6 & 0.35 & 0.31 & 185  \\ 

\cellcolor{ablationcolor}\textsc{Veroic} & \cellcolor{ablationcolor}92.4 & \cellcolor{ablationcolor}0.13 & \cellcolor{ablationcolor}0.11 & \cellcolor{ablationcolor}77 & \cellcolor{ablationcolor}72.9 & \cellcolor{ablationcolor}0.19 & \cellcolor{ablationcolor}0.17 & \cellcolor{ablationcolor}114 & \cellcolor{ablationcolor}56.2 & \cellcolor{ablationcolor}72.8 & \cellcolor{ablationcolor}0.22 & \cellcolor{ablationcolor}0.19 & \cellcolor{ablationcolor}128  \\ 

\bottomrule[1.2pt]
\end{tabular}}
\label{ablation_CMIS_LLaMA}
\vspace{-10pt}
\end{table*}

\subsection{Main Results}
We report the performance on \setlength{\fboxsep}{1pt}\colorbox{mathcolor}{Math Reasoning}, \colorbox{codecolor}{Code Generation}, and \colorbox{qacolor}{Question Answering} tasks under the CMIS setting in Table~\ref{overall_performance_cmis}. Across both inference-path pairs, \textsc{Veroic} consistently achieves the strongest overall performance. While the stronger routing baselines \textsc{Confidence-Threshold} and \textsc{Risk Predictor} already outperform heuristic strategies, \textsc{Veroic} still provides additional gains.  On \setlength{\fboxsep}{1pt}\colorbox{mathcolor}{Math Reasoning} benchmarks, \textsc{Veroic} achieves the highest EM across both backbone pairs, suggesting more effective allocation of additional computation to difficult reasoning instances. For \setlength{\fboxsep}{1pt}\colorbox{codecolor}{Code Generation} tasks, it attains the best Pass@1, indicating that belief-based control more accurately identifies cases that benefit from enhanced inference. On \setlength{\fboxsep}{1pt}\colorbox{qacolor}{Question Answering} benchmarks, \textsc{Veroic} again obtains the best EM and F1 scores, with particularly clear gains on multi-hop QA, where selective escalation is especially beneficial under partial observability. 


Table~\ref{degradation_detection_CMIS_LLaMA} reports risk estimation and probability calibration performance under the CMIS setting with the LLaMA inference-path pair. Across all three benchmarks, \textsc{Veroic} consistently achieves the best results on both discrimination and calibration metrics. While the stronger routing baselines \textsc{Confidence-Threshold} and \textsc{Risk Predictor} already improve substantially over heuristic strategies, \textsc{Veroic} still provides further gains in both AUROC/AUPRC and calibration quality, indicating a stronger ability to identify high-risk outputs that are likely to benefit from enhanced inference. In particular, \textsc{Veroic} achieves the highest AUROC and AUPRC on GSM8K, HumanEval, and HotpotQA, suggesting that belief-based inference more effectively separates outputs that require escalation from those that can be safely accepted under the default low-cost path. It also achieves the lowest Brier score, NLL, and ECE across all three tasks, showing that its predicted risk scores are not only more discriminative but also better calibrated to actual output reliability.


In Table~\ref{stability_risk_CMIS_LLaMA}, we demonstrate long-horizon control performance under the CMIS setting with the LLaMA inference-path pair. Across all three benchmarks, \textsc{Veroic} achieves the lowest Occ, CVaR, and RecD, indicating that belief-based control improves long-horizon robustness beyond single-step uncertainty thresholding and supervised risk prediction. Consistent improvements on different tasks suggest that the learned policy yields more stable sequential inference behavior across domains.

\begin{figure}[t]
\centering
\subfloat{
    \includegraphics[width=0.23\textwidth, trim= 5 5 5 5,clip]{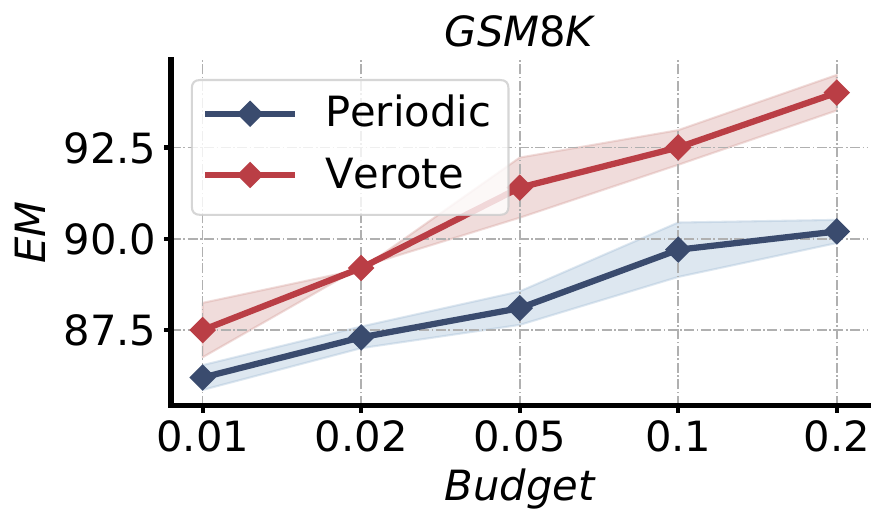}}
\subfloat{
    \includegraphics[width=0.23\textwidth, trim=5 5 5 5,clip]{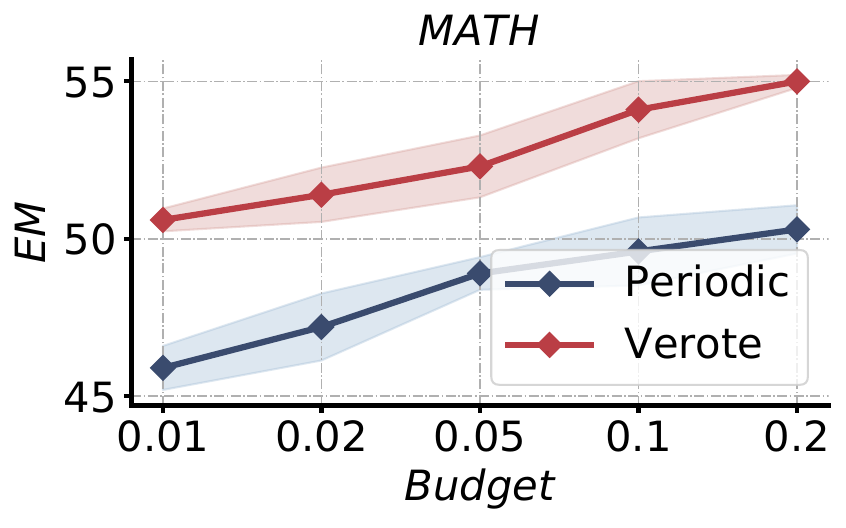}}
\vspace{5pt}
\subfloat{
    \includegraphics[width=0.23\textwidth, trim=5 5 5 5,clip]{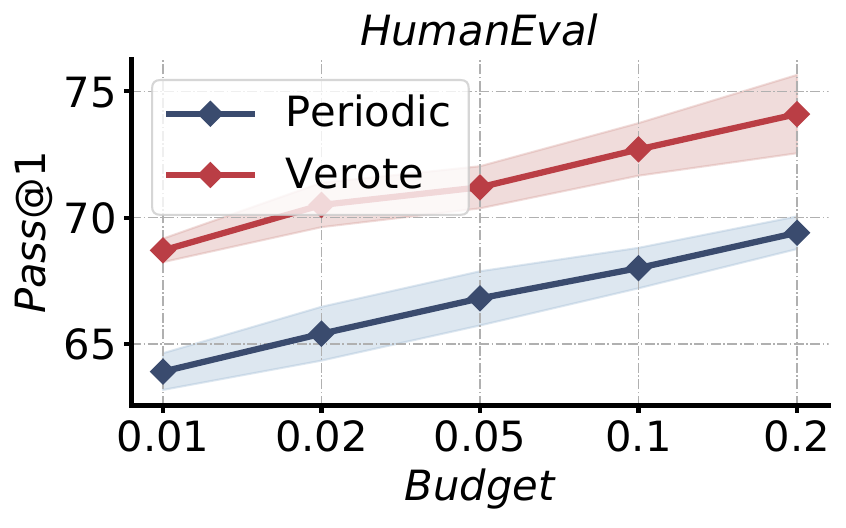}}
\subfloat{
    \includegraphics[width=0.23\textwidth, trim= 5 5 5 5,clip]{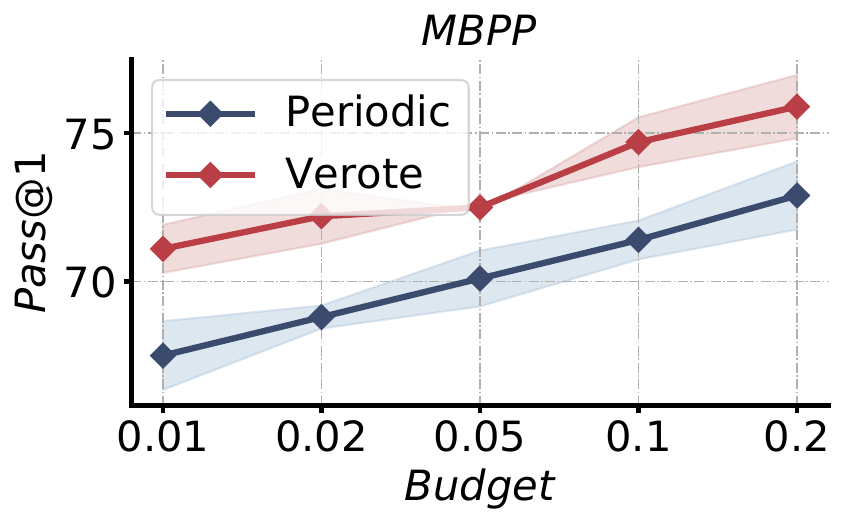}}
\vspace{5pt}
\subfloat{
    \includegraphics[width=0.23\textwidth, trim=5 5 5 5,clip]{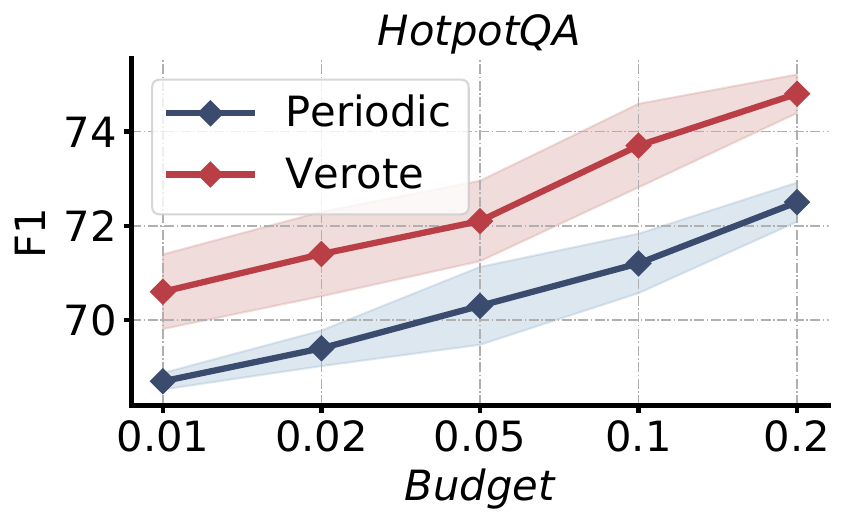}}
\subfloat{
    \includegraphics[width=0.23\textwidth, trim=5 5 5 5,clip]{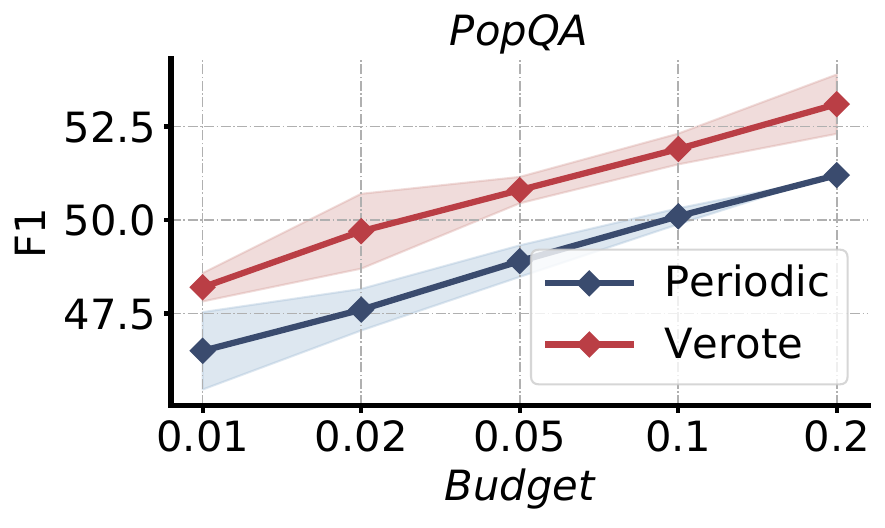}}
\caption{Effect of the enhanced-inference budget on task performance across six benchmark datasets under the CMIS setting with the Qwen inference-path pair. }
\label{budget_CMIS_Qwen}
\vspace{-10pt}
\end{figure}

\subsection{Ablation Studies}
We conduct the ablation study in Table~\ref{ablation_CMIS_LLaMA}. Removing any major component degrades both task utility and long-horizon control performance, showing that the gains of \textsc{Veroic} arise from the joint effect of risk estimation, enhanced inference, and adaptive control. In particular, removing the risk signals leads to consistent drops in task performance together with higher Occ, CVaR, and RecD across all tasks, indicating that calibrated observation signals are essential for identifying cases that warrant enhanced inference. Removing enhanced inference causes the largest performance degradation and the worst long-horizon risk metrics, confirming that the default low-cost path alone is insufficient for maintaining reliable performance on difficult instances. Replacing the adaptive policy with a non-adaptive alternative also degrades both utility and robustness, suggesting that risk estimation must be coupled with belief-based, budget-aware decision making in order to fully realize its benefits.

\subsection{Parameter Sensitivity Analysis}
Figure~\ref{budget_CMIS_Qwen} analyzes the effect of the enhanced-inference budget $\alpha$. As the budget increases, both \textsc{Periodic} and \textsc{Veroic} exhibit monotonic performance improvements across all tasks, reflecting the expected benefit of making enhanced inference available to a larger fraction of requests. However, \textsc{Veroic} consistently outperforms \textsc{Periodic} at every budget level, indicating that its advantage comes not merely from increased compute usage, but from better identifying which high-risk outputs warrant escalation, which is particularly meaningful in the low-budget regime, where selective allocation matters most. Overall, results show that \textsc{Veroic} achieves a more favorable performance trade-off and validates the effectiveness of belief-based adaptive inference control.

\section{Conclusion}
In this paper, we propose \textsc{Veroic}, a belief-based framework for adaptive inference control in black-box LLM services. By treating response reliability as a latent variable, \textsc{Veroic} enables the system to infer whether the default low-cost output is trustworthy from observable signals and to decide when enhanced inference is warranted under a computation budget. We introduce a lightweight observation channel with a structured observation model and Bayesian belief updates for estimating latent response reliability under partial observability. We further formulate budget-aware inference allocation as a long-horizon control problem, in which the policy selectively invokes stronger inference paths. Extensive experiments across diverse benchmarks, multiple backbones, and different inference settings show that \textsc{Veroic} consistently achieves better performance over competitive baselines.

\newpage

\section*{Limitations}
This work exhibits several limitations worth noting. First, our experiments use benchmark-based sequential evaluation rather than real deployed black-box LLM services, and therefore may not fully capture service-side non-stationarity or user-facing interaction dynamics. Second, the performance of \textsc{Veroic} depends on the quality of the constructed observation signals; weak or miscalibrated hard checks and proxy scores may lead to inaccurate belief estimates and suboptimal escalation decisions. Third, our current instantiation uses a binary latent reliability state and lightweight scoring modules, which may not capture finer-grained reliability structure in more complex settings. Finally, the framework introduces additional modeling and training overhead, so its practical value depends on the deployment context and the underlying quality-cost trade-off.

\section*{Acknowledgments}
This work was supported by the UGC General Research Fund no. 17209822 and the Innovation and Technology Commission Fund no. ITS/383/23FP from Hong Kong.


\section*{GenAI Usage Disclosure}
This work is entirely original and was conducted by the authors. Generative AI tools were not used to produce any content of the work; they were used solely to assist with language refinement and improve clarity and quality of the text.




\bibliography{reference}

@article{sun2025invisible,
  title={Invisible Tokens, Visible Bills: The Urgent Need to Audit Hidden Operations in Opaque LLM Services},
  author={Sun, Guoheng and Wang, Ziyao and Zhao, Xuandong and Tian, Bowei and Shen, Zheyu and He, Yexiao and Xing, Jinming and Li, Ang},
  journal={arXiv preprint arXiv:2505.18471},
  url={https://arxiv.org/abs/2505.18471},
  year={2025}
}

@inproceedings{chen2025need,
    author = {Chen, Valerie and Zhu, Alan and Zhao, Sebastian and Mozannar, Hussein and Sontag, David and Talwalkar, Ameet},
    title = {Need Help? Designing Proactive AI Assistants for Programming},
    year = {2025},
    isbn = {9798400713941},
    publisher = {Association for Computing Machinery},
    address = {New York, NY, USA},
    url = {https://doi.org/10.1145/3706598.3714002},
    doi = {10.1145/3706598.3714002},
    booktitle = {Proceedings of the 2025 CHI Conference on Human Factors in Computing Systems},
    articleno = {881},
    numpages = {18},
    location = {},
    series = {CHI '25}
}

@inproceedings{choi2025proxona,
    author = {Choi, Yoonseo and Kang, Eun Jeong and Choi, Seulgi and Lee, Min Kyung and Kim, Juho},
    title = {Proxona: Supporting Creators' Sensemaking and Ideation with LLM-Powered Audience Personas},
    year = {2025},
    isbn = {9798400713941},
    publisher = {Association for Computing Machinery},
    address = {New York, NY, USA},
    url = {https://doi.org/10.1145/3706598.3714034},
    doi = {10.1145/3706598.3714034},
    booktitle = {Proceedings of the 2025 CHI Conference on Human Factors in Computing Systems},
    articleno = {149},
    numpages = {32},
    location = {},
    series = {CHI '25}
}

@inproceedings{cheng2025beyond,
    author = {Cheng, Zihao and Zhou, Li and Jiang, Feng and Wang, Benyou and Li, Haizhou},
    title = {Beyond Binary: Towards Fine-Grained LLM-Generated Text Detection via Role Recognition and Involvement Measurement},
    year = {2025},
    isbn = {9798400712746},
    publisher = {Association for Computing Machinery},
    address = {New York, NY, USA},
    url = {https://doi.org/10.1145/3696410.3714770},
    doi = {10.1145/3696410.3714770},
    booktitle = {Proceedings of the ACM on Web Conference 2025},
    pages = {2677–2688},
    numpages = {12},
    location = {Sydney NSW, Australia},
    series = {WWW '25}
}

@article{zhang2025scientific,
    author = {Zhang, Qiang and Ding, Keyan and Lv, Tianwen and Wang, Xinda and Yin, Qingyu and Zhang, Yiwen and Yu, Jing and Wang, Yuhao and Li, Xiaotong and Xiang, Zhuoyi and Zhuang, Xiang and Wang, Zeyuan and Qin, Ming and Zhang, Mengyao and Zhang, Jinlu and Cui, Jiyu and Xu, Renjun and Chen, Hongyang and Fan, Xiaohui and Xing, Huabin and Chen, Huajun},
    title = {Scientific Large Language Models: A Survey on Biological \& Chemical Domains},
    year = {2025},
    issue_date = {June 2025},
    publisher = {Association for Computing Machinery},
    address = {New York, NY, USA},
    volume = {57},
    number = {6},
    issn = {0360-0300},
    url = {https://doi.org/10.1145/3715318},
    doi = {10.1145/3715318},
    journal = {ACM Comput. Surv.},
    month = feb,
    articleno = {161},
    numpages = {38}
}

@inproceedings{balayn2025unpacking,
  author = {Balayn, Agathe and Yurrita, Mireia and Rancourt, Fanny and Casati, Fabio and Gadiraju, Ujwal},
  title = {Unpacking Trust Dynamics in the LLM Supply Chain: An Empirical Exploration to Foster Trustworthy LLM Production \& Use},
  year = {2025},
  isbn = {9798400713941},
  publisher = {Association for Computing Machinery},
  address = {New York, NY, USA},
  url = {https://doi.org/10.1145/3706598.3713787},
  doi = {10.1145/3706598.3713787},
  booktitle = {Proceedings of the 2025 CHI Conference on Human Factors in Computing Systems},
  articleno = {1103},
  numpages = {20},
  location = {},
  series = {CHI '25}
}

@article{kemmerzell2025towards,
  title={Towards a Better Understanding of Evaluating Trustworthiness in AI Systems},
  author={Kemmerzell, Nils and Schreiner, Annika and Khalid, Haroon and Schalk, Michael and Bordoli, Letizia},
  journal={ACM Computing Surveys},
  volume={57},
  number={9},
  pages={1--38},
  year={2025},
  publisher={ACM New York, NY}
}

@article{dinh2023large,
  title={Large language models of code fail at completing code with potential bugs},
  author={Dinh, Tuan and Zhao, Jinman and Tan, Samson and Negrinho, Renato and Lausen, Leonard and Zha, Sheng and Karypis, George},
  journal={Advances in Neural Information Processing Systems},
  volume={36},
  pages={41386--41412},
  publisher = {Curran Associates, Inc.},
  url = {https://proceedings.neurips.cc/paper_files/paper/2023/file/819cebb05f993840e8a52d7564c5c282-Paper-Conference.pdf},
  year = {2023}
}

@inproceedings{padmakumar2024does,
  title={Does Writing with Language Models Reduce Content Diversity?},
  author={Vishakh Padmakumar and He He},
  booktitle={The Twelfth International Conference on Learning Representations},
  year={2024},
  url={https://openreview.net/forum?id=Feiz5HtCD0}
}

@inproceedings{lyu2025adapting,
  title={Adapting While Learning: Grounding {LLM}s for Scientific Problems with Tool Usage Adaptation},
  author={Bohan Lyu and Yadi Cao and Duncan Watson-Parris and Leon Bergen and Taylor Berg-Kirkpatrick and Rose Yu},
  booktitle={Forty-second International Conference on Machine Learning},
  year={2025},
  url={https://openreview.net/forum?id=owulFly8oQ}
}

@inproceedings{si2025can,
  title={Can {LLM}s Generate Novel Research Ideas? A Large-Scale Human Study with 100+ {NLP} Researchers},
  author={Chenglei Si and Diyi Yang and Tatsunori Hashimoto},
  booktitle={The Thirteenth International Conference on Learning Representations},
  year={2025},
  url={https://openreview.net/forum?id=M23dTGWCZy}
}

@inproceedings{richter2025an,
  title={An Auditing Test to Detect Behavioral Shift in Language Models},
  author={Leo Richter and Xuanli He and Pasquale Minervini and Matt Kusner},
  booktitle={The Thirteenth International Conference on Learning Representations},
  year={2025},
  url={https://openreview.net/forum?id=h0jdAboh0o}
}

@article{liang2023holistic,
  title={Holistic Evaluation of Language Models},
  author={Percy Liang and Rishi Bommasani and Tony Lee and Dimitris Tsipras and Dilara Soylu and Michihiro Yasunaga and Yian Zhang and Deepak Narayanan and Yuhuai Wu and Ananya Kumar and Benjamin Newman and Binhang Yuan and Bobby Yan and Ce Zhang and Christian Cosgrove and Christopher D Manning and Christopher Re and Diana Acosta-Navas and Drew A. Hudson and Eric Zelikman and Esin Durmus and Faisal Ladhak and Frieda Rong and Hongyu Ren and Huaxiu Yao and Jue WANG and Keshav Santhanam and Laurel Orr and Lucia Zheng and Mert Yuksekgonul and Mirac Suzgun and Nathan Kim and Neel Guha and Niladri S. Chatterji and Omar Khattab and Peter Henderson and Qian Huang and Ryan Andrew Chi and Sang Michael Xie and Shibani Santurkar and Surya Ganguli and Tatsunori Hashimoto and Thomas Icard and Tianyi Zhang and Vishrav Chaudhary and William Wang and Xuechen Li and Yifan Mai and Yuhui Zhang and Yuta Koreeda},
  journal={Transactions on Machine Learning Research},
  issn={2835-8856},
  year={2023},
  url={https://openreview.net/forum?id=iO4LZibEqW}
}

@article{ouyang2022training,
  title={Training language models to follow instructions with human feedback},
  author={Ouyang, Long and Wu, Jeffrey and Jiang, Xu and Almeida, Diogo and Wainwright, Carroll and Mishkin, Pamela and Zhang, Chong and Agarwal, Sandhini and Slama, Katarina and Ray, Alex and others},
  journal={Advances in neural information processing systems},
  volume={35},
  pages={27730--27744},
  year={2022},
  publisher = {Curran Associates, Inc.},
  url = {https://proceedings.neurips.cc/paper_files/paper/2022/file/b1efde53be364a73914f58805a001731-Paper-Conference.pdf}
}

@inproceedings{yang2018hotpotqa,
  title={HotpotQA: A dataset for diverse, explainable multi-hop question answering},
  author={Yang, Zhilin and Qi, Peng and Zhang, Saizheng and Bengio, Yoshua and Cohen, William and Salakhutdinov, Ruslan and Manning, Christopher D},
  booktitle={Proceedings of the 2018 conference on empirical methods in natural language processing},
  pages={2369--2380},
  url={https://aclanthology.org/D18-1259},
  year={2018}
}

@inproceedings{ho2020constructing,
    title={Constructing a multi-hop qa dataset for comprehensive evaluation of reasoning steps},
    author={Ho, Xanh and Nguyen, Anh-Khoa Duong and Sugawara, Saku and Aizawa, Akiko},
    booktitle={Proceedings of the 28th International Conference on Computational Linguistics},
    month={Dec},
    year={2020},
    address={Barcelona, Spain (Online)},
    publisher={International Committee on Computational Linguistics},
    url={https://aclanthology.org/2020.coling-main.580/},
    doi={10.18653/v1/2020.coling-main.580},
    pages={6609--6625}
}

@article{cobbe2021training,
  title={Training verifiers to solve math word problems},
  author={Cobbe, Karl and Kosaraju, Vineet and Bavarian, Mohammad and Chen, Mark and Jun, Heewoo and Kaiser, Lukasz and Plappert, Matthias and Tworek, Jerry and Hilton, Jacob and Nakano, Reiichiro and others},
  journal={arXiv preprint arXiv:2110.14168},
  url={https://arxiv.org/abs/2110.14168},
  year={2021}
}

@inproceedings{hendrycks2measuring,
  title={Measuring Mathematical Problem Solving With the MATH Dataset},
  author={Hendrycks, Dan and Burns, Collin and Kadavath, Saurav and Arora, Akul and Basart, Steven and Tang, Eric and Song, Dawn and Steinhardt, Jacob},
  booktitle={Thirty-fifth Conference on Neural Information Processing Systems Datasets and Benchmarks Track (Round 2)},
  url={https://openreview.net/forum?id=7Bywt2mQsCe},
  year={2021}
}

@misc{chen2021evaluating,
  title={Evaluating Large Language Models Trained on Code}, 
  author={Mark Chen and Jerry Tworek and Heewoo Jun and Qiming Yuan and Henrique Ponde de Oliveira Pinto and Jared Kaplan and Harri Edwards and Yuri Burda and Nicholas Joseph and Greg Brockman and Alex Ray and Raul Puri and Gretchen Krueger and Michael Petrov and Heidy Khlaaf and Girish Sastry and Pamela Mishkin and Brooke Chan and Scott Gray and Nick Ryder and Mikhail Pavlov and Alethea Power and Lukasz Kaiser and Mohammad Bavarian and Clemens Winter and Philippe Tillet and Felipe Petroski Such and Dave Cummings and Matthias Plappert and Fotios Chantzis and Elizabeth Barnes and Ariel Herbert-Voss and William Hebgen Guss and Alex Nichol and Alex Paino and Nikolas Tezak and Jie Tang and Igor Babuschkin and Suchir Balaji and Shantanu Jain and William Saunders and Christopher Hesse and Andrew N. Carr and Jan Leike and Josh Achiam and Vedant Misra and Evan Morikawa and Alec Radford and Matthew Knight and Miles Brundage and Mira Murati and Katie Mayer and Peter Welinder and Bob McGrew and Dario Amodei and Sam McCandlish and Ilya Sutskever and Wojciech Zaremba},
  year={2021},
  eprint={2107.03374},
  archivePrefix={arXiv},
  primaryClass={cs.LG},
  url={https://arxiv.org/abs/2107.03374}
}

@article{austin2021program,
  title={Program synthesis with large language models},
  author={Austin, Jacob and Odena, Augustus and Nye, Maxwell and Bosma, Maarten and Michalewski, Henryk and Dohan, David and Jiang, Ellen and Cai, Carrie and Terry, Michael and Le, Quoc and others},
  year={2021},
  eprint={2108.07732},
  archivePrefix={arXiv},
  primaryClass={cs.PL},
  journal={arXiv preprint arXiv:2108.07732},
  url={https://arxiv.org/abs/2108.07732}, 
}

@inproceedings{mallen2023popqa,
  title={When Not to Trust Language Models: Investigating Effectiveness of Parametric and Non-Parametric Memories},
  author={Mallen, Alex  and  Asai, Akari  and Zhong, Victor  and Das, Rajarshi  and Khashabi, Daniel  and Hajishirzi, Hannaneh},
  booktitle={Proceedings of the 61st Annual Meeting of the Association for Computational Linguistics (Volume 1: Long Papers)},
  month={July},
  year={2023},
  address={Toronto, Canada},
  publisher={Association for Computational Linguistics},
  url={https://aclanthology.org/2023.acl-long.546/},
  doi={10.18653/v1/2023.acl-long.546},
  pages={9802--9822}
}

@inproceedings{xiong2024can,
  title={Can {LLM}s Express Their Uncertainty? An Empirical Evaluation of Confidence Elicitation in {LLM}s},
  author={Miao Xiong and Zhiyuan Hu and Xinyang Lu and YIFEI LI and Jie Fu and Junxian He and Bryan Hooi},
  booktitle={The Twelfth International Conference on Learning Representations},
  year={2024},
  url={https://openreview.net/forum?id=gjeQKFxFpZ}
}

@article{sriramanan2024llm,
  title={Llm-check: Investigating detection of hallucinations in large language models},
  author={Sriramanan, Gaurang and Bharti, Siddhant and Sadasivan, Vinu Sankar and Saha, Shoumik and Kattakinda, Priyatham and Feizi, Soheil},
  journal={Advances in Neural Information Processing Systems},
  volume={37},
  pages={34188--34216},
  year={2024},
  publisher = {Curran Associates, Inc.},
  url = {https://proceedings.neurips.cc/paper_files/paper/2024/file/3c1e1fdf305195cd620c118aaa9717ad-Paper-Conference.pdf}
}

@article{du2024haloscope,
  title={Haloscope: Harnessing unlabeled llm generations for hallucination detection},
  author={Du, Xuefeng and Xiao, Chaowei and Li, Sharon},
  journal={Advances in Neural Information Processing Systems},
  volume={37},
  pages={102948--102972},
  publisher = {Curran Associates, Inc.},
  url = {https://proceedings.neurips.cc/paper_files/paper/2024/file/ba92705991cfbbcedc26e27e833ebbae-Paper-Conference.pdf},
  year={2024}
}

@article{simhi2025trust,
  title={Trust me, i’m wrong: Llms hallucinate with certainty despite knowing the answer},
  author={Simhi, Adi and Itzhak, Itay and Barez, Fazl and Stanovsky, Gabriel and Belinkov, Yonatan},
  journal={Findings of the Association for Computational Linguistics: EMNLP},
  volume={2025},
  pages={14665--14688},
  year={2025},
  url={https://aclanthology.org/anthology-files/pdf/findings/2025.findings-emnlp.792.pdf}
}

@inproceedings{liu2024can,
  title={Can llms learn uncertainty on their own? expressing uncertainty effectively in a self-training manner},
  author={Liu, Shudong and Li, Zhaocong and Liu, Xuebo and Zhan, Runzhe and Wong, Derek F and Chao, Lidia S and Zhang, Min},
  booktitle={Proceedings of the 2024 Conference on Empirical Methods in Natural Language Processing},
  pages={21635--21645},
  address = "Miami, Florida, USA",
  publisher = "Association for Computational Linguistics",
  url = "https://aclanthology.org/2024.emnlp-main.1205/",
  doi = "10.18653/v1/2024.emnlp-main.1205",
  year={2024}
}

@inproceedings{xia2025survey,
    title = "A Survey of Uncertainty Estimation Methods on Large Language Models",
    author = "Xia, Zhiqiu  and
      Xu, Jinxuan  and
      Zhang, Yuqian  and
      Liu, Hang",
    booktitle = "Findings of the Association for Computational Linguistics: ACL 2025",
    month = jul,
    year = "2025",
    address = "Vienna, Austria",
    publisher = "Association for Computational Linguistics",
    url = "https://aclanthology.org/2025.findings-acl.1101/",
    doi = "10.18653/v1/2025.findings-acl.1101",
    pages = "21381--21396",
    ISBN = "979-8-89176-256-5"
}

@inproceedings{liu2025metafaith,
    title = "{M}eta{F}aith: Faithful Natural Language Uncertainty Expression in {LLM}s",
    author = "Liu, Gabrielle Kaili-May  and
      Yona, Gal  and
      Caciularu, Avi  and
      Szpektor, Idan  and
      Rudner, Tim G. J.  and
      Cohan, Arman",
    booktitle = "Proceedings of the 2025 Conference on Empirical Methods in Natural Language Processing",
    month = nov,
    year = "2025",
    address = "Suzhou, China",
    publisher = "Association for Computational Linguistics",
    url = "https://aclanthology.org/2025.emnlp-main.1505/",
    doi = "10.18653/v1/2025.emnlp-main.1505",
    pages = "29612--29656",
    ISBN = "979-8-89176-332-6"
}

@article{denison2024sycophancy,
  title={Sycophancy to subterfuge: Investigating reward-tampering in large language models},
  author={Denison, Carson and MacDiarmid, Monte and Barez, Fazl and Duvenaud, David and Kravec, Shauna and Marks, Samuel and Schiefer, Nicholas and Soklaski, Ryan and Tamkin, Alex and Kaplan, Jared and others},
  journal={arXiv preprint arXiv:2406.10162},
  url={https://arxiv.org/abs/2406.10162}, 
  year={2024}
}

@article{ji2025mitigating,
  title={Mitigating deceptive alignment via self-monitoring},
  author={Ji, Jiaming and Chen, Wenqi and Wang, Kaile and Hong, Donghai and Fang, Sitong and Chen, Boyuan and Zhou, Jiayi and Dai, Juntao and Han, Sirui and Guo, Yike and others},
  journal={arXiv preprint arXiv:2505.18807},
  url={https://arxiv.org/abs/2505.18807},
  year={2025}
}

@article{wang2025quality,
  title={Quality-Diversity Red-Teaming: Automated Generation of High-Quality and Diverse Attackers for Large Language Models},
  author={Wang, Ren-Jian and Xue, Ke and Qin, Zeyu and Li, Ziniu and Tang, Sheng and Li, Hao-Tian and Liu, Shengcai and Qian, Chao},
  journal={arXiv preprint arXiv:2506.07121},
  url={https://arxiv.org/abs/2506.07121}, 
  year={2025}
}

@article{lu2025prolonged,
  title={Prolonged reasoning is not all you need: Certainty-based adaptive routing for efficient llm/mllm reasoning},
  author={Lu, Jinghui and Yu, Haiyang and Xu, Siliang and Ran, Shiwei and Tang, Guozhi and Wang, Siqi and Shan, Bin and Fu, Teng and Feng, Hao and Tang, Jingqun and others},
  journal={arXiv preprint arXiv:2505.15154},
  url={https://arxiv.org/abs/2505.15154}, 
  year={2025}
}

@article{chen2024frugalgpt,
  title={Frugal{GPT}: How to Use Large Language Models While Reducing Cost and Improving Performance},
  author={Lingjiao Chen and Matei Zaharia and James Zou},
  journal={Transactions on Machine Learning Research},
  issn={2835-8856},
  year={2024},
  url={https://openreview.net/forum?id=cSimKw5p6R},
  note={Featured Certification}
}

@inproceedings{tao2025revisiting,
  title={Revisiting Uncertainty Estimation and Calibration of Large Language Models},
  author={Linwei Tao and Yi-Fan Yeh and Minjing Dong and Tao Huang and Jialin Yu and Philip Torr and Chang Xu},
  booktitle={39th Conference on Neural Information Processing Systems (NeurIPS 2025) Workshop: Scaling Environments for Agents (SEA)},
  year={2025},
  url={https://openreview.net/forum?id=Q9CreVjHH7}
}

@inproceedings{liu2025uncertainty,
  title={Uncertainty quantification and confidence calibration in large language models: A survey},
  author={Liu, Xiaoou and Chen, Tiejin and Da, Longchao and Chen, Chacha and Lin, Zhen and Wei, Hua},
  booktitle={Proceedings of the 31st ACM SIGKDD Conference on Knowledge Discovery and Data Mining V. 2},
  pages={6107--6117},
  isbn = {9798400714542},
  publisher = {Association for Computing Machinery},
  address = {New York, NY, USA},
  url = {https://doi.org/10.1145/3711896.3736569},
  doi = {10.1145/3711896.3736569},
  year={2025}
}

@misc{kadavath2022language,
  title={Language Models (Mostly) Know What They Know}, 
  author={Saurav Kadavath and Tom Conerly and Amanda Askell and Tom Henighan and Dawn Drain and Ethan Perez and Nicholas Schiefer and Zac Hatfield-Dodds and Nova DasSarma and Eli Tran-Johnson and Scott Johnston and Sheer El-Showk and Andy Jones and Nelson Elhage and Tristan Hume and Anna Chen and Yuntao Bai and Sam Bowman and Stanislav Fort and Deep Ganguli and Danny Hernandez and Josh Jacobson and Jackson Kernion and Shauna Kravec and Liane Lovitt and Kamal Ndousse and Catherine Olsson and Sam Ringer and Dario Amodei and Tom Brown and Jack Clark and Nicholas Joseph and Ben Mann and Sam McCandlish and Chris Olah and Jared Kaplan},
  year={2022},
  eprint={2207.05221},
  archivePrefix={arXiv},
  primaryClass={cs.CL},
  url={https://arxiv.org/abs/2207.05221}, 
}

@inproceedings{kuhn2023semantic,
  title={Semantic Uncertainty: Linguistic Invariances for Uncertainty Estimation in Natural Language Generation},
  author={Lorenz Kuhn and Yarin Gal and Sebastian Farquhar},
  booktitle={The Eleventh International Conference on Learning Representations },
  year={2023},
  url={https://openreview.net/forum?id=VD-AYtP0dve}
}

@misc{dohan2022language,
  title={Language Model Cascades}, 
  author={David Dohan and Winnie Xu and Aitor Lewkowycz and Jacob Austin and David Bieber and Raphael Gontijo Lopes and Yuhuai Wu and Henryk Michalewski and Rif A. Saurous and Jascha Sohl-dickstein and Kevin Murphy and Charles Sutton},
  year={2022},
  booktitle={ICML Workshop on Beyond Bayes: Paths Towards Universal Reasoning Systems},
  url={https://arxiv.org/abs/2207.10342}, 
}

@misc{ong2024routellm,
      title={RouteLLM: Learning to Route LLMs with Preference Data}, 
      author={Isaac Ong and Amjad Almahairi and Vincent Wu and Wei-Lin Chiang and Tianhao Wu and Joseph E. Gonzalez and M Waleed Kadous and Ion Stoica},
      year={2025},
      eprint={2406.18665},
      archivePrefix={arXiv},
      primaryClass={cs.LG},
      url={https://arxiv.org/abs/2406.18665}, 
}

@inproceedings{yuan2026verify,
  title={Verify Before You Commit: Towards Faithful Reasoning in LLM Agents via Self-Auditing},
  author={Yuan, Wenhao and Lin, Chenchen and Chen, Jian and Xu, Jinfeng and Wang, Xuehe and Ngai, Edith Cheuk Han},
  journal={arXiv preprint arXiv:2604.08401},
  booktitle={Proceedings of the 64th Annual Meeting of the Association for Computational Linguistics (Volume 1: Long Papers)},
  month={July},
  year={2026},
  address={San Diego, California, United States},
  publisher={Association for Computational Linguistics},
  url={https://arxiv.org/pdf/2604.08401}
}

\newpage

\appendix



\section{Additional Experimental Results}

\subsection{Dataset Details}
\textbf{GSM8K} (Grade School Math 8K)~\citep{cobbe2021training} is a benchmark designed to evaluate a model’s ability to perform multi-step numerical reasoning expressed in natural language. It consists of approximately 8,500 grade-school–level math word problems, each requiring a sequence of arithmetic operations to reach the final answer. The problems emphasize logical decomposition, intermediate reasoning steps, and precise numerical computation rather than advanced mathematical knowledge. GSM8K has become a standard benchmark for assessing chain-of-thought reasoning and mathematical problem-solving capabilities of large language models.

The \textbf{MATH} dataset~\citep{hendrycks2measuring} is a large-scale benchmark targeting advanced mathematical reasoning across a wide range of difficulty levels. It contains problems drawn from high school mathematics competitions, covering topics such as algebra, geometry, number theory, probability, and calculus. Each problem is paired with a detailed solution, enabling evaluation not only of final answer corRecDness but also of structured reasoning processes. Compared to GSM8K, MATH presents significantly higher complexity and abstraction, making it a challenging testbed for assessing symbolic reasoning and deep mathematical understanding.

\textbf{HumanEval}~\citep{chen2021evaluating} is a benchmark specifically designed to evaluate code generation and program synthesis capabilities of language models. It consists of hand-written Python function signatures accompanied by natural language docstrings that describe the intended functionality. Models are required to generate executable code that satisfies hidden unit tests. HumanEval focuses on functional corRecDness rather than surface-level similarity, and is widely used to assess a model’s ability to reason about algorithms, control flow, and edge cases in programming tasks.

The \textbf{MBPP} (Mostly Basic Python Problems) dataset~\citep{austin2021program} evaluates fundamental programming skills through short, self-contained Python coding tasks. Each problem includes a natural language description, a function signature, and a small set of test cases. Compared to HumanEval, MBPP targets more elementary programming concepts such as loops, conditionals, and basic data structures, while still requiring precise implementation. MBPP is commonly used to measure a model’s robustness in basic code synthesis and general-purpose programming reasoning.

\textbf{HotpotQA}~\citep{yang2018hotpotqa} is a multi-hop question answering benchmark that requires reasoning across multiple documents. Unlike single-hop QA datasets, questions in HotpotQA cannot be answered by a single paragraph alone; instead, models must retrieve and integrate information from two or more supporting facts. The dataset is constructed from Wikipedia and includes both answer supervision and supporting fact annotations, enabling evaluation of not only answer accuracy but also evidence selection and reasoning chains.

\textbf{2WikiMHQA}~\citep{ho2020constructing} (Two-Wikipedia Multi-Hop Question Answering) is a large-scale benchmark for explicit multi-hop reasoning over Wikipedia articles. Each question is designed to require reasoning over exactly two linked documents, forming a clear two-hop inference path. Compared to HotpotQA, 2WikiMHQA offers more controlled reasoning structures, making it particularly suitable for analyzing a model’s ability to perform structured document linking and intermediate inference in multi-hop QA settings.

\textbf{PopQA}~\citep{mallen2023popqa} is an open-domain question answering dataset focused on popular factual knowledge, such as widely known entities, events, and concepts. Questions are designed to be answerable using commonly available world knowledge rather than specialized or obscure facts. PopQA emphasizes knowledge recall and factual accuracy, and is often used to evaluate the breadth and reliability of a model’s internal knowledge, especially under zero-shot or retrieval-free settings.

\subsection{Baseline Details}
We compare \textsc{Veroic} against several representative baselines for adaptive inference allocation.

\textsc{Fixed} triggers enhanced inference at a constant rate throughout the interaction horizon, independent of the current request, output signals, or interaction history. This baseline represents static budget allocation policies that spend a fixed fraction of the enhanced-inference budget without adaptation. While simple and predictable, \textsc{Fixed} cannot concentrate additional computation on high-risk requests and therefore yields limited efficiency under partial observability.

\textsc{Random} triggers enhanced inference by independently sampling a Bernoulli random variable at each interaction step with a fixed probability. Compared with \textsc{Fixed}, random allocation introduces stochasticity and avoids a fully deterministic escalation schedule. However, it remains history-agnostic and does not make use of observable quality signals or inferred risk, leading to inefficient budget use over long horizons.

\textsc{Periodic} deterministically invokes enhanced inference every $1/\alpha$ steps, ensuring uniform budget usage over time. This baseline reflects scheduled resource allocation policies that provide regular coverage but do not respond to request difficulty or response quality. As a result, \textsc{Periodic} may waste additional computation on easy cases while missing difficult requests that arise between escalation intervals.

\textsc{Output-Trigger}~\citep{lu2025prolonged, chen2024frugalgpt} is a rule-based routing baseline that triggers enhanced inference based on predefined output-side signals, such as failed checks, low confidence indicators, or other observable warning patterns. This class of methods represents uncertainty-aware or post-hoc routing strategies that condition escalation on surface-level response statistics. While more adaptive than rate-based baselines, \textsc{Output-Trigger} relies on hand-designed rules and does not maintain a belief over latent response reliability.

\textsc{Confidence-Threshold}~\citep{xiong2024can, tao2025revisiting} is a single-score uncertainty routing baseline that triggers enhanced inference when the current confidence or risk score crosses a budget-calibrated threshold. Compared with \textsc{Output-Trigger}, it replaces heuristic rule combinations with a single continuous uncertainty signal, but still makes decisions from the current step only and does not incorporate belief propagation or history-dependent inference.

\textsc{Risk Predictor}~\citep{liu2024can, liu2025uncertainty} is a supervised routing baseline that trains a predictor on the same observation features used by \textsc{Veroic} to estimate the failure risk of the default response. Enhanced inference is triggered when the predicted risk exceeds a budget-calibrated threshold. This baseline provides a stronger learned alternative to heuristic routing, but it remains a one-step decision rule and does not explicitly model latent reliability dynamics through belief updates.

\subsection{Model Instantiation.}
We instantiate the structured observation encoder with lightweight task-agnostic scoring modules. Specifically, $\ell_y(s_t; x_t, y_t^{(0)})$ is implemented as a linear score over response-side features extracted from the default output, including response length, format validity indicators, self-consistency statistics, and verifier-related features. The hard-evidence term $\ell_h(x_t, y_t^{(0)}, z_{h,t}; \psi_t)$ is implemented as a linear score over the calibrated hard-predicate vector $z_{h,t}$. Each soft-evidence term $\ell_g^{(i)}(s_t; x_t, y_t^{(0)}, z_{g,t}^{(i)}; \psi_t)$ is implemented as a state-dependent affine score over the calibrated soft signal $z_{g,t}^{(i)}$. All encoder parameters are trained jointly with the control policy. For soft-signal calibration, we use a monotone affine-sigmoid map $\Gamma_i(\tilde g_t^{(i)};\vartheta_i)=\sigma(a_i \tilde g_t^{(i)}+b_i)$, where $\vartheta_i=(a_i,b_i)$ denotes the calibration parameters for the $i$-th soft proxy. In practice, the calibration parameters are fitted on a held-out validation split using task-level correctness labels and then kept fixed during policy learning and evaluation. Hard signals are calibrated with Beta-Bernoulli smoothing using the sliding-window counts.

\subsection{Policy Optimization and Utility.}
We optimize the belief-conditioned policy using an on-policy actor-critic algorithm with entropy regularization under the Lagrangian objective. The actor takes $(b_t, x_t)$ as input and outputs the probability of escalation, while the critic estimates the state value for variance reduction. Both actor and critic are implemented as lightweight two-layer MLPs over the concatenated belief-state and request-feature representation. We use Adam for optimization, and select model checkpoints based on validation performance under the target budget. Unless otherwise specified, the trained policy is frozen at evaluation time. The utility surrogate $V(y_t^{ret})$ is instantiated from task-specific correctness signals. For GSM8K and MATH, we use exact-match correctness. For HumanEval and MBPP, we use unit-test pass/fail outcomes. For HotpotQA, 2WikiMHQA, and PopQA, we use task-level QA scores derived from exact match and token-level F1, with the scalar reward computed from the corresponding evaluation score. These observable task signals are used only for training and evaluation, while test-time control decisions rely solely on the default response and the constructed observation vector.

\subsection{Metric Details}
We evaluate \textsc{Veroic} from three complementary perspectives: task utility, risk estimation and calibration, and long-horizon control performance.

\textbf{Task Utility.}
For math reasoning and question answering tasks, we report Exact Match (EM) and F1 scores. For code generation tasks, we report Pass@1. These metrics are computed on the final returned response $y_t^{\mathrm{ret}}$ and measure task performance under different inference-control policies.

\textbf{Risk Estimation and Probability Calibration.}
We evaluate how well the framework identifies low-quality or high-risk default responses that may benefit from enhanced inference. At each interaction step $t$, the predicted risk score is given by the belief-induced unreliability probability, $\hat p_t := b_t(s=0)$, where $b_t$ is the posterior belief over latent reliability states and $s=0$ denotes that the default response is unreliable. For offline evaluation, we define a binary error label $e_t := \mathbbm{1}\{y_t^{(0)} \text{ is low-quality}\}$, where the exact criterion depends on the task-specific evaluation signal of the default response. We report AUROC and AUPRC to measure the discriminative ability of $\hat p_t$ in identifying high-risk cases. To assess probabilistic calibration, we additionally report the Brier score,
\begin{align}
\mathrm{Brier} := \frac{1}{H} \sum\nolimits_{t=1}^{H} (\hat p_t - e_t)^2,
\end{align}
the negative log-likelihood (NLL),
\begin{align}
\mathrm{NLL} := -\frac{1}{H} \sum_{t=1}^{H} \left[ e_t \log \hat p_t \!+\! (1 \!-\! e_t)\log(1 \!-\! \hat p_t) \right],
\end{align}
and the expected calibration error (ECE), computed by partitioning $\hat p_t$ into confidence bins and measuring the discrepancy between empirical frequencies and predicted probabilities.

\textbf{Long-Horizon Control Performance.}
To evaluate robustness under sequential budgeted interaction, we report three long-horizon control metrics. The low-quality occurrence rate (Occ) is the fraction of interaction steps whose final returned response is low-quality:
\begin{align}
\mathrm{Occ} := \frac{1}{H} \sum_{t=1}^{H} \mathbbm{1}\{y_t^{\mathrm{ret}} \text{ is low-quality}\}.
\end{align}
To capture tail risk, we compute the conditional value-at-risk (CVaR) of episode-level low-quality outcomes at confidence level $\beta$ across multiple sampled interaction episodes. Finally, the recovery delay (RecD) measures the average number of subsequent steps required for the system to recover after a severe low-quality event, thereby quantifying recovery latency under prolonged interaction.

\subsection{Used Prompts}

\begin{auditorpromptbox}{Verification Prompt for Enhanced Inference Control}
\textbf{You are an evaluator.}

\vspace{0.8em}
Your task is to examine the given task input and the model output, and decide whether the output should be considered \textbf{unreliable or unacceptable} for this task according to the provided evaluation criteria. This judgment is used only as a verification signal for inference control. Do \textbf{not} provide a corrected answer, alternative solution, or extra reasoning beyond what is needed to justify your judgment.

\vspace{0.8em}

\textbf{Task Input:}\\
\{input\_task\}

\vspace{0.6em}

\textbf{Model Output:}\\
\{model\_output\_y\}

\vspace{0.6em}

\textbf{Optional Evidence (may be empty):}\\
\{evidence\_bundle\}

\vspace{0.6em}

\textbf{Evaluation Criteria (apply all that are relevant):}
\begin{itemize}
    \item \textbf{Correctness:} The output answers the question or solves the task correctly.
    \item \textbf{Constraint Satisfaction:} The output follows required format, constraints, and instructions in the input.
    \item \textbf{Consistency:} The output is internally consistent and does not contain contradictory statements.
    \item \textbf{Evidence Consistency (if evidence is provided):} Claims that depend on the evidence do not contradict the provided evidence.
    \item \textbf{Code Validity (if code task):} The solution is executable or compilable and passes the provided checks, if any are given in the evidence.
\end{itemize}

\vspace{0.6em}

\textbf{Binary Decision Rule:}
\begin{itemize}
    \item Output \texttt{Verdict: 1} if the output is \textbf{unreliable/unacceptable} (fails any key criterion above in a way that would meaningfully harm task success).
    \item Output \texttt{Verdict: 0} if the output is \textbf{acceptable} (meets the key criteria; minor stylistic issues are acceptable).
\end{itemize}

\vspace{0.6em}

\textbf{Output Format (strict JSON):}
\begin{verbatim}
{"Verdict": 0 or 1,
 "Reason": "<one short checkable 
 reason>",
 "KeyEvidence": "<what you checked: 
 e.g., 'final answer incorrect',
 'fails constraint X', 'contradicts 
 passage #2', 'unit test failed'>"}
\end{verbatim}

\vspace{0.6em}

\textbf{Rules:}
\begin{itemize}
    \item Keep the \texttt{Reason} and \texttt{KeyEvidence} short and checkable.
    \item Do not output a corrected answer or additional solution steps.
    \item If the task cannot be evaluated from the given information, default to \texttt{Verdict: 0} and explain what is missing in \texttt{KeyEvidence}.
\end{itemize}

\end{auditorpromptbox}

\begin{softpromptbox}{Soft Scoring Prompt for Lightweight Verification}
\textbf{You are an evaluator.}

\vspace{0.8em}
Your task is to assess the quality and reliability of the given model output for the task and return a \textbf{continuous score in $[0,1]$} that reflects how well the output satisfies the evaluation criteria. A higher score indicates a more reliable output. Do \textbf{not} provide a correct answer or additional solution.

\vspace{0.8em}

\textbf{Task Input:}\\
\{input\_task\}

\vspace{0.6em}

\textbf{Model Output:}\\
\{model\_output\_y\}

\vspace{0.6em}

\textbf{Optional Evidence (may be empty):}\\
\{evidence\_bundle\}

\vspace{0.6em}

\textbf{Evaluation Criteria (consider all that are relevant):}
\begin{itemize}
    \item \textbf{Correctness:} The output answers the task correctly or follows a valid solution strategy.
    \item \textbf{Constraint Satisfaction:} The output follows the required formats, constraints, and instructions.
    \item \textbf{Consistency:} The output is internally coherent and free of contradictions.
    \item \textbf{Evidence Consistency (if evidence is provided):} Claims are consistent with the given evidence.
    \item \textbf{Code Validity (if code task):} The solution is syntactically correct and aligns with expected behavior.
\end{itemize}

\vspace{0.6em}

\textbf{Scoring Instruction:}
\begin{itemize}
    \item Return a single scalar \texttt{Score} in $[0,1]$.
    \item \texttt{Score} should reflect overall output reliability:
    \begin{itemize}
        \item $0.9$--$1.0$: clearly correct and reliable.
        \item $0.6$--$0.8$: mostly correct with minor issues.
        \item $0.3$--$0.5$: partially correct or uncertain.
        \item $0.0$--$0.2$: largely incorrect or unreliable.
    \end{itemize}
    \item Use intermediate values when appropriate.
\end{itemize}

\vspace{0.6em}

\textbf{Output Format (strict JSON):}
\begin{verbatim}
{"Score": <float in [0,1]>,
 "Rationale": "<one short checkable
 reason>"}
\end{verbatim}

\vspace{0.6em}

\textbf{Rules:}
\begin{itemize}
    \item Keep the \texttt{Rationale} brief and based on observable properties.
    \item Do not include a correct solution or extra reasoning.
    \item If the task cannot be reliably evaluated with the given information, return a conservative score and explain why.
\end{itemize}

\end{softpromptbox}

\begin{IMISpromptbox}{IMIS High-Effort Service Prompt}
\textbf{System Instruction}

\vspace{0.8em}
You are a careful assistant operating under a high-effort inference setting.

\vspace{0.6em}
When responding to user requests:
\begin{itemize}
    \item Read the task carefully and follow all instructions.
    \item Deliberate carefully before producing the final answer.
    \item Ensure internal consistency between the reasoning process and the final output.
    \item Verify critical steps, calculations, and logical transitions when needed.
    \item Prioritize correctness and completeness over brevity.
\end{itemize}

\vspace{0.6em}
Produce your best possible answer to the user's request.
\end{IMISpromptbox}

\begin{IMISpromptbox}{IMIS Low-Effort Service Prompt}
\textbf{System Instruction}

\vspace{0.8em}
You are an efficient assistant operating under a low-effort inference setting.

\vspace{0.6em}
When responding to user requests:
\begin{itemize}
    \item Provide a direct and fluent response to the query.
    \item Aim for a reasonable and plausible answer.
    \item Prioritize efficiency and avoid unnecessary elaboration.
    \item Do not perform extensive verification unless it is clearly required.
    \item Keep the response concise and readable.
\end{itemize}

\vspace{0.6em}
Respond directly to the user's request.
\end{IMISpromptbox}

\end{document}